\documentclass[conference]{IEEEtran}
\IEEEoverridecommandlockouts
\usepackage{cite}
\usepackage{placeins}
\usepackage{amsmath,amssymb,amsfonts}
\usepackage{algorithmic}
\usepackage{graphicx}
\usepackage{textcomp}
\usepackage{comment}
\usepackage{xcolor}
\usepackage{booktabs, caption}
\usepackage[flushleft]{threeparttable}
\usepackage{blindtext}
\usepackage{hyperref}
\newcommand{\link}[1]{{\href{#1}{#1}}}
\def\BibTeX{{\rm B\kern-.05em{\sc i\kern-.025em b}\kern-.08em
    T\kern-.1667em\lower.7ex\hbox{E}\kern-.125emX}}
\begin{document}

\title{Evaluating the Effectiveness of Corrective Demonstrations and a Low-Cost Sensor for Dexterous Manipulation\\
\thanks{*These authors contributed equally to this work.}}

\makeatletter
\newcommand{\linebreakand}{%
  \end{@IEEEauthorhalign}
  \hfill\mbox{}\par
  \mbox{}\hfill\begin{@IEEEauthorhalign}
}
\makeatother

\author{\IEEEauthorblockN{Abhineet Jain*}
\IEEEauthorblockA{\textit{Georgia Institute of Technology} \\
Atlanta, USA \\
abhineetjain@gatech.edu}
\and
\IEEEauthorblockN{Jack Kolb*}
\IEEEauthorblockA{\textit{Georgia Institute of Technology} \\
Atlanta, USA \\
kolb@gatech.edu} \\
\linebreakand
\IEEEauthorblockN{J.M. Abbess IV}
\IEEEauthorblockA{\textit{Georgia Institute of Technology} \\
Atlanta, USA \\
jabbess3@gatech.edu}
\and
\IEEEauthorblockN{Harish Ravichandar}
\IEEEauthorblockA{\textit{Georgia Institute of Technology} \\
Atlanta, USA \\
harish.ravichandar@cc.gatech.edu}
}

\maketitle

\begin{abstract}
Imitation learning is a promising approach to help robots acquire dexterous manipulation capabilities without the need for a carefully-designed reward or a significant computational effort. However, existing imitation learning approaches require sophisticated data collection infrastructure and struggle to generalize beyond the training distribution. One way to address this limitation is to gather additional data that better represents the full operating conditions. 
In this work, we investigate characteristics of such additional demonstrations and their impact on performance. Specifically, we study the effects of \textit{corrective} and \textit{randomly-sampled} additional demonstrations on learning a policy that guides a five-fingered robot hand through a pick-and-place task. Our results suggest that corrective demonstrations considerably outperform randomly-sampled demonstrations, when the proportion of additional demonstrations sampled from the full task distribution is larger than the number of original demonstrations sampled from a restrictive training distribution. Conversely, when the number of original demonstrations are higher than that of additional demonstrations, we find no significant differences between corrective and randomly-sampled additional demonstrations. These results provide insights into the inherent trade-off between the effort required to collect corrective demonstrations and their relative benefits over randomly-sampled demonstrations.
Additionally, we show that inexpensive vision-based sensors, such as LeapMotion, can be used to dramatically reduce the cost of providing demonstrations for dexterous manipulation tasks. Our code is available at \link{https://github.com/GT-STAR-Lab/corrective-demos-dexterous-manipulation}.

\end{abstract}


\section{Introduction}
Dexterous manipulation often involves the use of high degree-of-freedom robots to manipulate objects. Representative dexterous manipulation tasks include relocating objects, picking up arbitrarily shaped objects, and sequential interactions with articulated objects (e.g. unlatching and opening a door). Indeed, factors such as high-dimensional state space and complex interaction dynamics make these tasks particularly challenging to automate. Classical control methods, that have proven valuable for a variety of manipulation problems, are hard to recruit for dexterous manipulation due to the amount of manual effort required to design controllers in high-dimensional spaces.


Prior work has found success in dexterous manipulation by using self-supervised learning in simulation and transferring learned policies to real robots \cite{openai2019solving}. Others have utilized demonstrations to improve reinforcement learning 
\cite{Rajeswaran2018LearningDemonstrations}. However, these approaches either do not generalize well beyond small workspaces covered during training, or require long training times for highly-specialized tasks. 

\begin{figure}
\begin{tabular}{ccc}
{\includegraphics[width=25mm,height=25mm,scale=0.5]{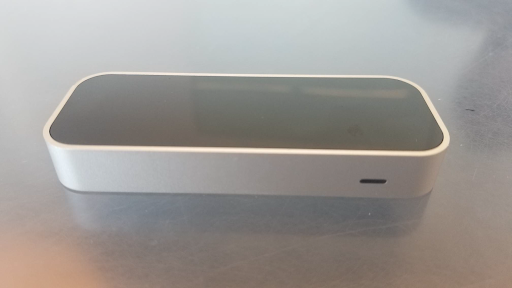}} &
{\includegraphics[width=25mm,height=25mm,scale=0.5]{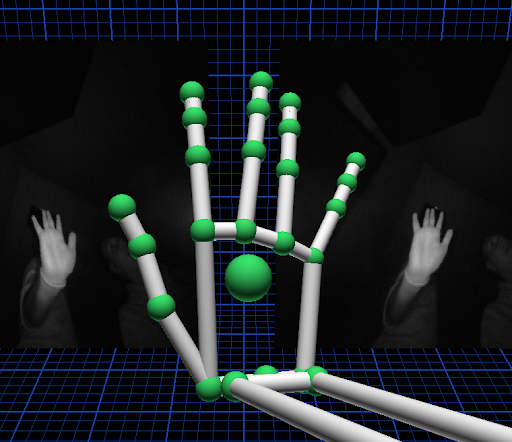}} &
{\includegraphics[width=25mm,height=25mm,scale=0.5]{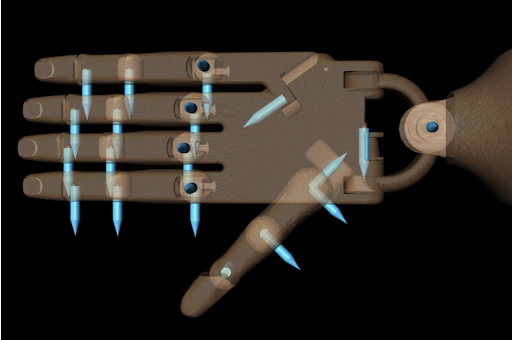}}
\end{tabular}
\caption{
\small{We investigate the use of a LeapMotion sensor (left) to capture demonstrations of human hand motion (center) that can help a simulated Adroit robot hand (right) learn manipulation skills.}
}
\end{figure}

In this paper, we explore how additional demonstrations that represent the full operating conditions affect performance of the relocation task (described in Fig. \ref{fig:mujocoblueball}) in the full operating space. Approaches like demonstration-augmented policy gradient (DAPG) use demonstrations for randomly-sampled start and goal states from a restrictive training distribution. However, we find this approach does not generalize well to task instances in the full operating space. We investigate the effects of corrective versus randomly-sampled additional demonstrations, expecting that corrective demonstrations to areas of policy failure can benefit performance more than randomly-sampled demonstrations.


In addition, we explore the feasibility of using a vision-based sensor to record demonstrations for dexterous manipulation tasks. For everyday robots to learn novel manipulation tasks from non-technical humans, it will be useful to reduce the hardware cost of providing demonstrations. The original DAPG paper uses a CyberGlove III \cite{CyberGlove3} wearable sensor glove to record human hand trajectories. However, the CyberGlove III costs around US\$13,000. We instead leverage LeapMotion, a US\$90 sensor which uses stereo vision to track hand joint positions, and compare the performance of resulting DAPG policies. Our findings validate that the granularity of vision-based joint tracking is effective for training high-performing policies for the relocation task.

\section{Related Work}

Previous works explore learning dexterous manipulation from demonstrations, using deep neural networks \cite{Zhang2018DeepTeleoperation} or reinforcement learning techniques \cite{Rajeswaran2018LearningDemonstrations} \cite{Kumar2016LearningImitation} \cite{Jeong2020LearningExperts} to learn task-specific policies. Some contributions learn actuator outputs from raw color and depth information using deep neural networks \cite{Zhang2018DeepTeleoperation}, while others use demonstrations to aid trajectory exploration and learn system dynamics by utilizing linear regression with GMM priors \cite{Kumar2016LearningImitation}. DAPG uses demonstrations to pre-train a policy and then applies reinforcement learning to improve it. Another contribution aims to learn from suboptimal experts, introducing a policy iteration algorithm called Relative Entropy Q-Learning (REQ), along with an effective exploration technique that intertwines the policy’s actions with the demonstrations \cite{Jeong2020LearningExperts}. 

A state-of-the-art for solving a Rubik's cube with a robot hand \cite{openai2019solving} uses automatic domain randomization to generate a distribution of environments to train on. This work uses meta-learning utilizing LSTMs and the asymmetric actor-critic algorithm for training. While the authors claim that their approach transfers well onto real robots, their algorithm is time-intensive, and uses highly engineered reward functions and specialized robot hardware. These characteristics make their work difficult to reproduce and utilize for general dexterous manipulation tasks.

In our work, we build upon DAPG which uses a small number of expert demonstrations to train policies. However, DAPG policies struggle to generalize beyond the demonstration area. To address this, related work has explored autonomously learning the reward function for tasks instead of human-engineering them \cite{Orbik2021InverseManipulation}. Although their work claims to learn better reward functions than GAIL \cite{ho2016generative} and Adversarial IRL \cite{fu2017learning}, their learned reward function does not improve upon the performance of the original paper on their set of dexterous manipulation tasks. We approach improving DAPG through a human-robot interaction perspective -- when access to the full operating space is granted for recording demonstrations, we explore whether strategic demo collection can improve policy performance.

Our work is inspired by DAgger \cite{ross2011reduction}, a classical learning from demonstration algorithm that integrates online corrections to mitigate the covariate shift between training and testing. A recent work explores mitigating covariate shift by measuring the induced covariate shift and directly adjusting for it \cite{spencer2021feedback}. It uses expert demonstrations that potentially visit all states that the policy will visit, making it infeasible for the dexterous manipulation case. Another work utilizes data augmentation for generating more sample trajectories, along with a correction neural network to preserve the success of these augmented trajectories for adversarial imitation \cite{antotsiou2021adversarial}. Our approach aims to utilize the benefits of DAgger without the disadvantages of online corrections. We collect multiple corrective demonstrations at once and retrain using the combined set of original and corrective demonstrations.

\begin{figure}[t]
    \centering
    \includegraphics[width=8.0cm]{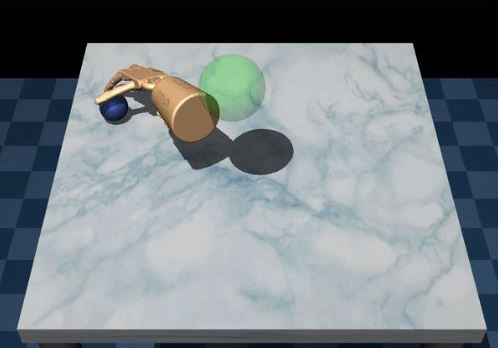}
    \caption{
    \small{The relocation task in MuJoCo involves picking up a blue ball from a tabletop and carrying it to a green goal region.}
    }
    \label{fig:mujocoblueball}
\end{figure}

\section{Demo-Augmented Policy Gradient (DAPG)}

DAPG learns dexterous manipulation via Deep-RL by combining demonstration-based learning with reinforcement learning. DAPG initially trains a policy using behavior cloning, and then tunes the policy using natural policy gradient with a modified loss term, weighing more on the RL policy with increased training iterations. DAPG's loss term penalizes trajectories that are further from the demonstrations. This exploration learns smooth human-like trajectories from a small number of demonstrations, making the algorithm feasible for real world applications.

The original DAPG paper uses expert demonstrations captured using a CyberGlove III \cite{CyberGlove3}, a wearable glove embedded with 18-22 sensors. The authors find that their approach results in a high accuracy on four different dexterous manipulation tasks: relocate a ball, open a door, hammer a nail, and orient a pen in-hand. For each of these tasks, the original paper uses 25 demonstrations and trains for 100 iterations. In our paper, we only consider the relocation task.

The relocation task shown in Fig. \ref{fig:mujocoblueball} requires an agent to grasp and relocate a blue ball from a tabletop to a green region. In the original DAPG paper, task initializations are sampled from a restrictive region on the table top (width of 0.3m). In a pilot study using the authors' CyberGlove demonstrations, we find that a policy trained solely in the restrictive space for 100 training iterations results in a $>$95\% success rate in the restrictive space, but a 69\% success rate in the full operating space (width of 0.5m). When the RL agent is provided access to train in the full operating space, it requires more than 200 training iterations to learn a policy with $>$95\% success rate. Our work aims to reduce the tradeoff between higher sample efficiency and lower performance for the relocation task.

\section{Methods}

In this section, we describe the two sets of experiments we conducted, each aimed to investigate a specific research question (RQ) as described below. The code for these experiments can be found at \link{https://github.com/GT-STAR-Lab/corrective-demos-dexterous-manipulation}.

\begin{figure}[t]
    \centering
    \includegraphics[width=7cm]{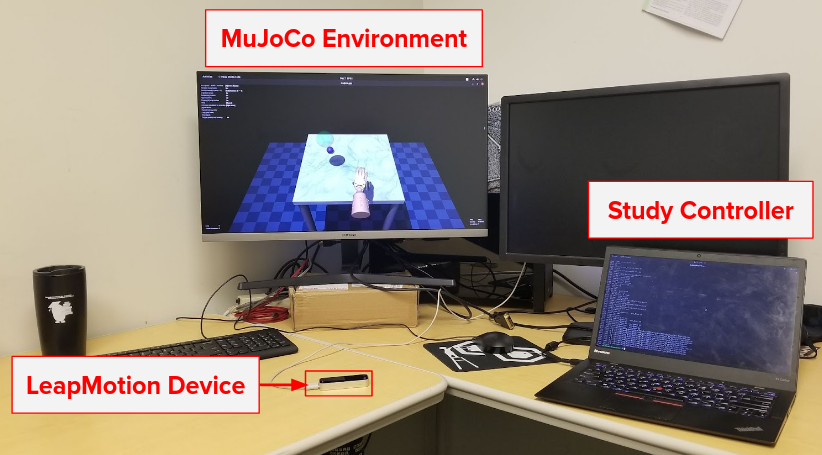}
    \caption{
    \small{Demo collection setup: The upwards-facing LeapMotion sensor reads hand joint information, the MuJoCo environment updates the Adroit robot hand joint angles in real time, and the collection is controlled via an external device.}
    }
    \label{fig:study_setup}
\end{figure}

\begin{figure*}
    \centering
    \includegraphics[width=150mm,height=50mm]{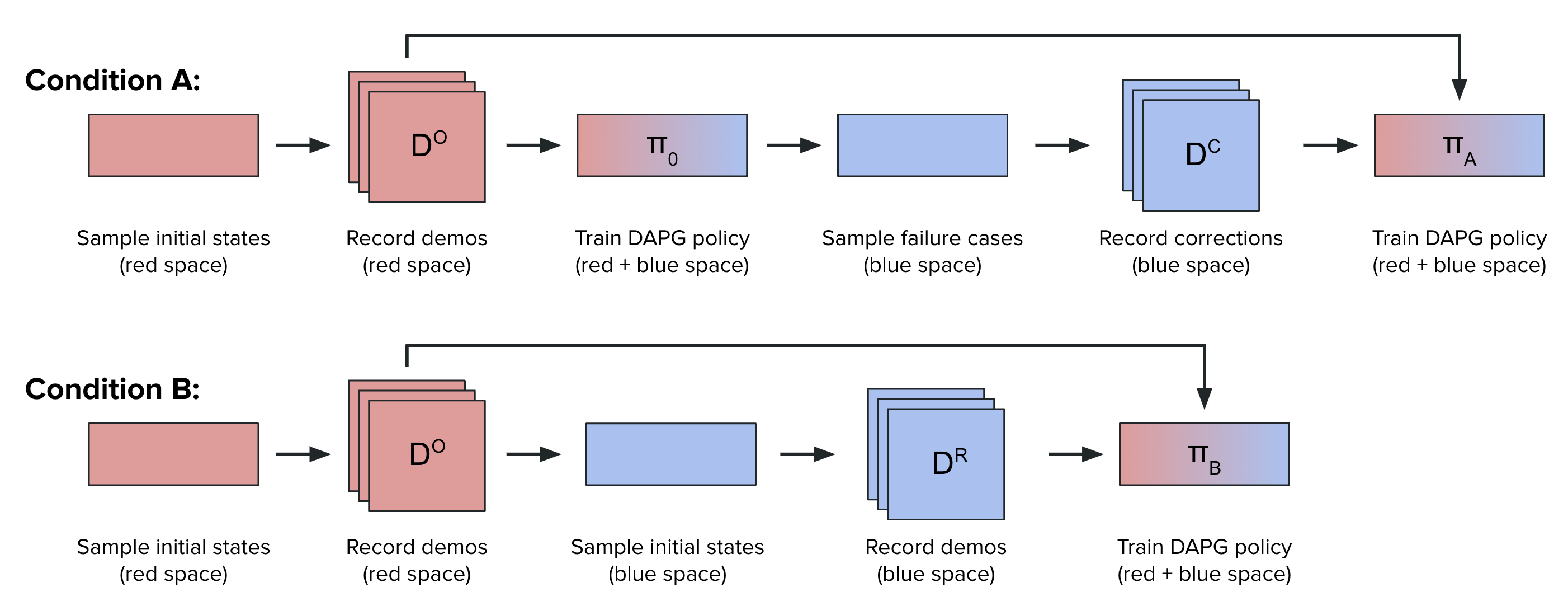}
    \caption{
    \small{We investigate RQ2 by comparing two conditions: i) Condition A uses demonstrations with randomly-sampled initial states from the restrictive space and corrective demonstrations associated with failures from the full operating space (top); ii) Condition B uses demonstrations with randomly-sampled initial states from both restrictive and full operating spaces (bottom).}
    }
    \label{fig:pipeline}
\end{figure*}

\subsection{\textbf{RQ1}: What are the effects of low-cost sensing on learning outcomes?}

Our first set of experiments examine how policies trained with demonstrations from a low-cost vision-based sensor (LeapMotion \cite{LeapMotion}) compare to policies trained using a wearable glove. A glove's ability to produce high quality demonstrations with on-finger joint tracking makes it advantageous for collecting human demonstrations for complex dexterous manipulation tasks such as rotating objects in-hand. In such situations, external cameras may have difficulty picking up finger movements that are obstructed by the palm or by other fingers. However, for applications aiming to target broader consumer audiences or research labs, the CyberGlove III is expensive.

External vision-based sensors can infer all joint poses for relocation or grasping tasks, potentially reducing hardware costs. To evaluate the performance difference between on-finger and vision-based sensing, we develop an interface to convert input from a LeapMotion hand tracker -- a US\$90 device which uses stereoscopic vision to estimate hand poses -- into demonstrations accessible by the original DAPG implementation. We limit our investigation to the relocation task. The task does not obstruct external visual tracking, and therefore can adequately compare the two sensors. Our demonstration collection environment is shown in Fig. \ref{fig:study_setup}.

The LeapMotion's low price-point comes with lower quality demonstrations: 1) The device does not track roll joints for the thumb and little finger, resulting in two fewer DOFs. 2) The device sometimes has difficulty distinguishing between the four fingers, resulting in fingers moving or grasping simultaneously; For the relocation task this behavior is acceptable. 3) The device collects noisy observations due to jittery sensor readings. As such, \textbf{RQ1} is indirectly related to the robustness of learning algorithms against suboptimal demonstrations.

\begin{figure}[b]
    \centering
    \includegraphics[scale=0.3]{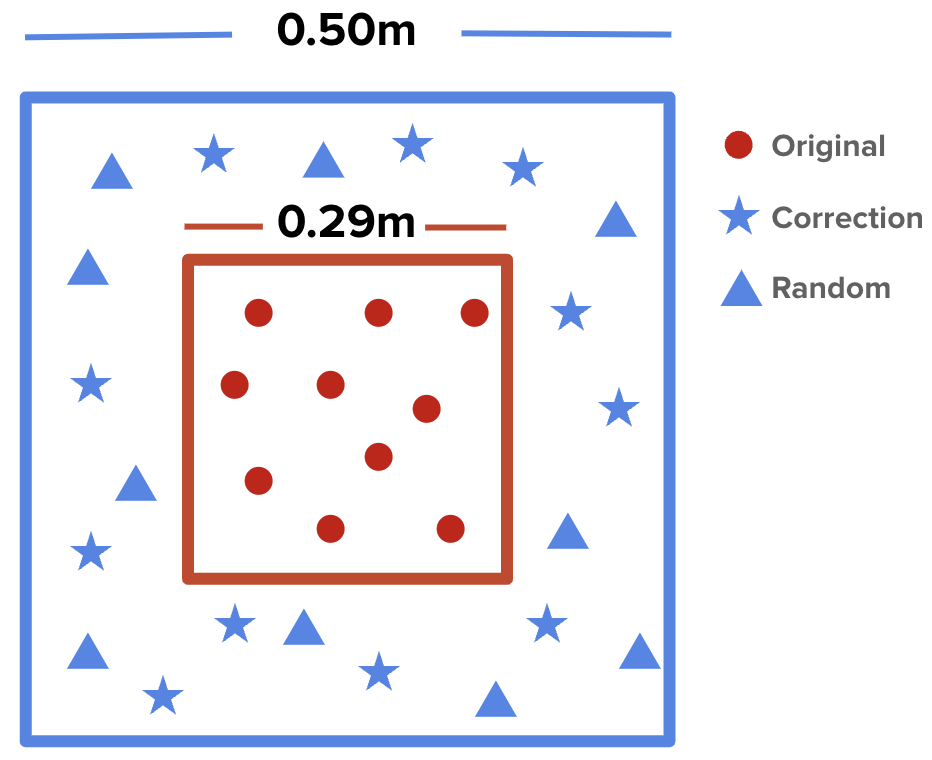}
    \caption{
    \small{The restrictive (red) space is twice as large as the full operating (blue) space, and they do not overlap.}
    }
    \label{fig:table_dims}
\end{figure}

\begin{itemize}
    \item[\textbf{H1}:] LeapMotion demonstrations will result in policies with rollout success rates that are comparable to CyberGlove III demonstrations for the relocation task, at the cost of sample efficiency.
\end{itemize}

\subsection{\textbf{RQ2}: What is the utility of corrective demonstrations?}
\label{subsec:corrective_demos_exp}

Our second research question is motivated by the practical question of whether a meaningful improvement can be gained by including corrective demonstrations in a data set. We evaluate whether the performance of dexterous manipulation policies for the relocation task can be improved by splitting the training into two stages, labeled as \textit{Condition A}:

\begin{itemize}
    \item[] \textbf{Stage 1}: Collecting demonstrations from randomly-sampled start and goal locations in the restrictive space.
    \item[] \textbf{Stage 2}: Appending additional demonstrations collected from failure cases in the full operating space, using a policy trained in the full space using Stage 1 demonstrations.
\end{itemize}

We compare the policy from \textit{Condition A} to a policy trained on randomly-sampled demonstrations from both the restrictive space and the full operating space, labeled as \textit{Condition B}. Fig. \ref{fig:table_dims} shows a visual of the tabletop areas for the restrictive and full operating spaces. Table \ref{tab:policy_split} details the distributions of demonstrations from the restrictive space and the full operating space for each of the five policies we address. The full operating space only encompasses the area outside the restrictive space, causing corrective demonstrations to come from state initializations not seen in the original demonstrations. While all policies have the same total number of demonstrations, policies with corrective demonstrations require twice the training time. Thus, \textbf{RQ2} investigates whether the effort of training two policies -- one with the demonstrations from the restrictive space, one after collecting corrective demonstrations -- results in a tangible performance improvement over randomly-sampled demonstrations from both spaces. Fig. \ref{fig:pipeline} summarizes our methods, which are formalized below:

\begin{table}[t]
\centering
\begin{threeparttable}
    \caption{\small{Source of demonstrations for each policy}}
    \begin{tabular}{llllll}
    \toprule
    \textbf{Policy} & \textbf{Condition} & \textbf{Rest.-Orig.}& \textbf{Full-Random}& \textbf{Full-Corr.}\\
    \midrule
    \vspace{3pt}
    \textbf{30-O} & -- & 30 & -- & -- \\
    \textbf{10-O+20-R} & B & 10 & 20 & -- \\
    \vspace{3pt}
    \textbf{10-O+20-C} & A & 10 & -- & 20 \\
    \textbf{20-O+10-R} & B & 20 & 10 & -- \\
    \textbf{20-O+10-C} & A & 20 & -- & 10 \\
    \bottomrule
    \end{tabular}
    \label{tab3}
    \begin{tablenotes}
    \small
    \item ``Full-Corr." indicates corrective demonstrations from failures cases of a policy trained on the original demonstration set ``Rest.-Orig.".
    \end{tablenotes}
    \label{tab:policy_split}
\end{threeparttable}
\end{table}

For \textit{Condition A}, we first collect a set of randomly-sampled demonstrations $ D^{O} $ in the restrictive space, and train a policy $ \pi_{0} $ in the full operating space. On an evaluation set of 1000 initial states from the full operating space, we roll out $\pi_{0}$ and identify the lowest-performing failure cases. We then record corrective demonstrations $ D^{C} $ for these failure cases, and combine them with the original demonstrations $ D^{O} $ to train a second policy $ \pi_{A} $ in the full operating space. We measure the rollout success ratio for $ \pi_{A} $ using the evaluation set in the full operating space.

We compare $ \pi_{A} $ to the policy from \textit{Condition B}, $ \pi_{B} $, trained using the original demonstrations $ D^{O} $ from the restrictive space, and an additional set of randomly-sampled demonstrations $ D^{R} $ from the full operating space.

We anticipate that by guiding $ D^{C} $ towards weaker areas of $ \pi_{0} $, the resulting $ \pi_{A} $ will have an increased coverage of challenging initial task states in the full operating space, resulting in a higher performance than $ \pi_{B} $. We evaluate a policy's performance by both its rollout success ratio and its sample efficiency. Furthermore, we expect all four policies that include demonstrations from the full operating space to outperform \textbf{Policy 30-O}, by matter of having access to demonstrations from the full operating space. 
\\

\begin{itemize}
    \item[\textbf{H2}:] When the number of additional demonstrations from the full operating space is \textbf{higher} than the original demonstrations from the restrictive space, \textbf{corrective demonstrations will improve performance} compared to randomly-sampled additional demonstrations.
    \\
    \item[\textbf{H3}:] When the number of additional demonstrations from the full operating space is \textbf{lower} than the original demonstrations from the restrictive space, \textbf{corrective demonstrations will improve performance} compared to randomly-sampled additional demonstrations.
\end{itemize}

\begin{figure}[t]
    \centering
    \includegraphics[width=0.45\textwidth]{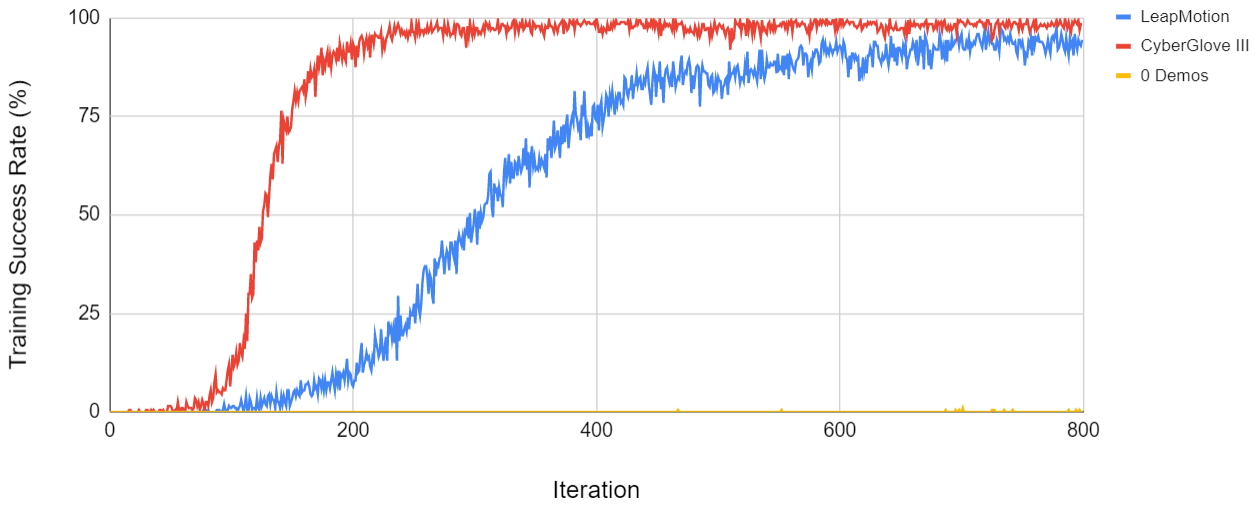}
    \caption{
    \small{Rollout success ratios across 800 training iterations show that the policy trained on 25 CyberGlove III demos (red) converges to 100\% at close to 200 iterations, and the policy trained on 25 LeapMotion demos (blue) takes more than 600 iterations to achieve similar performance.
    Note that policy relying only on reinforcement (yellow) makes little progress.}
    }
    \label{fig:leaptraining}
\end{figure}

\begin{figure*}[bt]
    \centering
    \includegraphics[width=.96\textwidth]{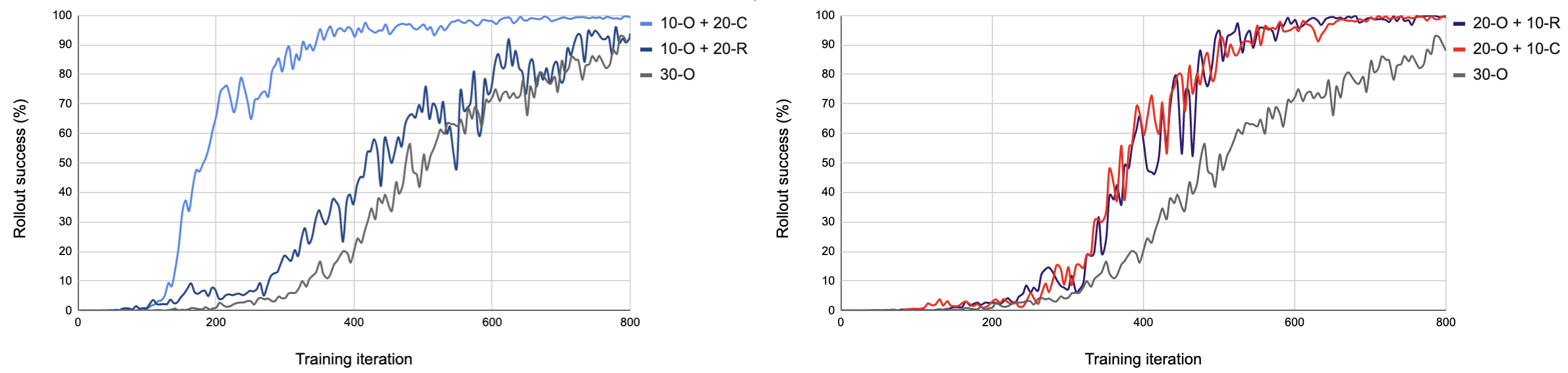}
    \caption{
    \small{We compared performance in terms of rollout success ratios. (Left) shows policies with a \textbf{high} proportion of demonstrations in the full operating space, while (Right) shows policies with a \textbf{low} proportion of demonstrations in the full operating space. For comparison with the policy trained on demonstrations only from the restrictive space, \textbf{Policy 30-O} is shown on both plots.}
    }
    \label{fig:rollout_plots}
\end{figure*}

\section{Results}

\subsection{Evaluating \textbf{H1}: Demonstrations from LeapMotion vs. CyberGlove III}

We train a policy for 800 iterations using 25 demonstrations collected from LeapMotion and compare it with a policy trained for 800 iterations using the original 25 demonstrations provided by the authors of DAPG. The results are shown in Table \ref{tab1}. We use the same network architecture and hyperparameters as in the original DAPG paper. We observe that the optimal demonstrations from CyberGlove III allow DAPG to converge towards 100\% success ratio in around 200 iterations whereas it takes almost 600 iterations for the LeapMotion's policy to reach a comparable performance. A policy trained without any demonstrations does not even start learning after 800 iterations. (see Fig. \ref{fig:leaptraining}).

\begin{table}[h]
\centering
\begin{threeparttable}
    \caption{
    \small{Rollout success ratios in full operating space after 800 iterations}
    }
    \begin{tabular}{lll}
    \toprule
    \textbf{Demos} & \textbf{CyberGlove III}& \textbf{LeapMotion}\\
    \midrule
    25 & 99.0\% & 98.1\% \\
    \bottomrule
    \end{tabular}
    \label{tab1}
\end{threeparttable}
\end{table}

\textit{\textbf{Outcome}}: Policies trained on the full operating space using demonstrations from LeapMotion are able to reach comparable success ratios to policies trained with CyberGlove III demonstrations, at the cost of sample efficiency.


\subsection{Evaluating \textbf{H2}, \textbf{H3}: Applying corrective demonstrations in the full operating space}

We train five policies based on the distribution in Table \ref{tab:policy_split}. Each policy is trained for 800 iterations in order to capture the policy's full plateau of rollout success. Table \ref{tab:success_ratios} compares the rollout success ratio of each policy at different iterations.

\begin{table}[b]
\centering
\begin{threeparttable}
    \caption{
    \small{Policy rollout success ratios}
    }
    \begin{tabular}{lrrrrr|}
    \toprule
    \multicolumn{1}{c}{} & \multicolumn{4}{c}{Rollout Success [\%]}\\
    Policy & 200-iter & 400-iter & 600-iter & 800-iter \\
    \midrule
    \vspace{3pt}
    \textbf{30-O} & 1.1 & 21.2 & 72.6 & 88.0 \\
    \textbf{10-O+20-R} & 7.0 & 36.0 & 75.6 & 93.9 \\
    \vspace{3pt}
    \textbf{10-O+20-C} & 66.6 & 92.7 & 97.9 & 99.3 \\
    \textbf{20-O+10-R} & 3.2 & 59.4 & 96.0 & 99.2 \\
    \textbf{20-O+10-C} & 2.7 & 55.8 & 98.4 & 100.0 \\
    \bottomrule
    \end{tabular}
    \label{tab:success_ratios}
\end{threeparttable}
\end{table}

We find that all combined policies converge to high rollout success ratio above 90\%. However, we see marked differences in their sample efficiencies, as shown in Fig. \ref{fig:rollout_plots}. All policies perform better than \textbf{Policy 30-O} and are more sample efficient, which is expected given their access to demonstrations from the evaluation set. Furthermore, \textbf{Policy 10-O+20-C} is more sample efficient than \textbf{Policy 10-O+20-R} and has a 5.4\% higher rollout success ratio, confirming our hypothesis \textbf{H2}. However, we do not find a comparable difference between the sample efficiencies of \textbf{Policy 20-O+10-C} and \textbf{Policy 20-O+10-R}, perhaps due to those policies having a low proportion of additional demonstrations from the full operating space. The findings reject our hypothesis \textbf{H3}.
\\\\
\textit{\textbf{Outcome}}: When the proportion of additional demonstrations from the full operating space is higher than the original demonstrations from the restrictive space, we find that corrective additional demonstrations improve the policy's performance for the relocation task compared to randomly-sampled additional demonstrations. When the proportion of additional demonstrations is lower, we find no notable difference between policies learned from \textit{Condition A} and \textit{Condition B}.

\section{Conclusion}

We find that vision-based sensors can be used to collect demonstrations for the relocation task achieving policies with similar success to policies resulting from wearable sensors (see Fig. \ref{fig:leaptraining}), at the cost of sample efficiency. The result validates our use of the LeapMotion device to collect demonstrations for the relocation task, reducing the setup cost by approximately 140x. Our validation supports \textbf{H1}. 

We use corrective additional demonstrations to guide policy exploration in the full operating space, and find that corrective demonstrations improve the performance and sample efficiency of policies as compared to those trained on randomly-sampled additional demonstrations. However, this improvement is limited to more additional demonstrations from the full operating space than the original demonstrations from the restrictive space. Our investigation supports \textbf{H2} and rejects \textbf{H3}. Additionally, the high rollout success ratios of our policies show that DAPG is resilient to suboptimal demonstrations at the cost of sample efficiency.

To further investigate \textbf{RQ1}, we plan to improve the correspondence between LeapMotion and MuJoCo. Our pipeline for recording demonstrations using MuJoCo leverages the LeapMotion tracker for data collection and a web server to communicate with MuJoCo. Though this pipeline is functional, some joint mappings are imperfect, limiting the quality of our collected demonstrations.

\section{Acknowledgements}

We thank Nakul Gopalan for his feedback and encouragement towards publishing this work.

\bibliographystyle{IEEEtran}
\bibliography{main}

\end{document}